\documentclass[10pt,twocolumn,letterpaper]{article}

\usepackage{authblk}

\usepackage[pagenumbers]{cvpr} 

\usepackage[dvipsnames]{xcolor}

\usepackage{adjustbox}
\usepackage{multirow}
\usepackage{caption}
\usepackage{subcaption}
\usepackage{amssymb}
\usepackage{pifont}
\usepackage[dvipsnames]{xcolor}
\newcommand{\cmark}{\textcolor{Green}{\text{\ding{51}}}}%
\newcommand{\xmark}{\textcolor{Red}{\text{\ding{55}}}}%

\usepackage{amsmath}

\DeclareMathOperator*{\argmin}{argmin}

\usepackage{algorithm}
\usepackage{algorithmic}

\usepackage{newfloat}
\usepackage{listings}
\DeclareCaptionStyle{ruled}{labelfont=normalfont,labelsep=colon,strut=off} 
\floatstyle{ruled}
\newfloat{listing}{tb}{lst}{}
\floatname{listing}{Listing}

\usepackage{etoolbox}
\makeatletter
\AfterEndEnvironment{algorithm}{\let\@algcomment\relax}
\AtEndEnvironment{algorithm}{\kern2pt\hrule\relax\vskip3pt\@algcomment}
\let\@algcomment\relax
\newcommand\algcomment[1]{\def\@algcomment{\footnotesize#1}}
\renewcommand\fs@ruled{\def\@fs@cfont{\bfseries}\let\@fs@capt\floatc@ruled
  \def\@fs@pre{\hrule height.8pt depth0pt \kern2pt}%
  \def\@fs@post{}%
  \def\@fs@mid{\kern2pt\hrule\kern2pt}%
  \let\@fs@iftopcapt\iftrue}
\makeatother

\definecolor{cvprblue}{rgb}{0.21,0.49,0.74}
\usepackage[pagebackref,breaklinks,colorlinks,citecolor=cvprblue]{hyperref}

\def\paperID{?} 
\def\confName{CVPR}
\def\confYear{2024}

\author[1,2]{\vspace{-0.4cm} Kam Woh Ng}
\author[1,3]{Xiatian Zhu}
\author[1,2]{Yi-Zhe Song}
\author[1,2]{Tao Xiang}

\affil[1]{CVSSP, University of Surrey}
\affil[2]{iFlyTek-Surrey Joint Research Centre}
\affil[3]{Surrey Institute for People-Centred AI}
\affil[ ]{\texttt{\small \{kamwoh.ng,xiatian.zhu,y.song,t.xiang\}@surrey.ac.uk}}

\title{DreamCreature: Crafting Photorealistic Virtual Creatures from Imagination}

\begin{document}

\twocolumn[{%
\renewcommand\twocolumn[1][]{#1}%
\maketitle

\begin{center}
    \centering
    \vspace{-0.4cm}
    \includegraphics[width=\textwidth]{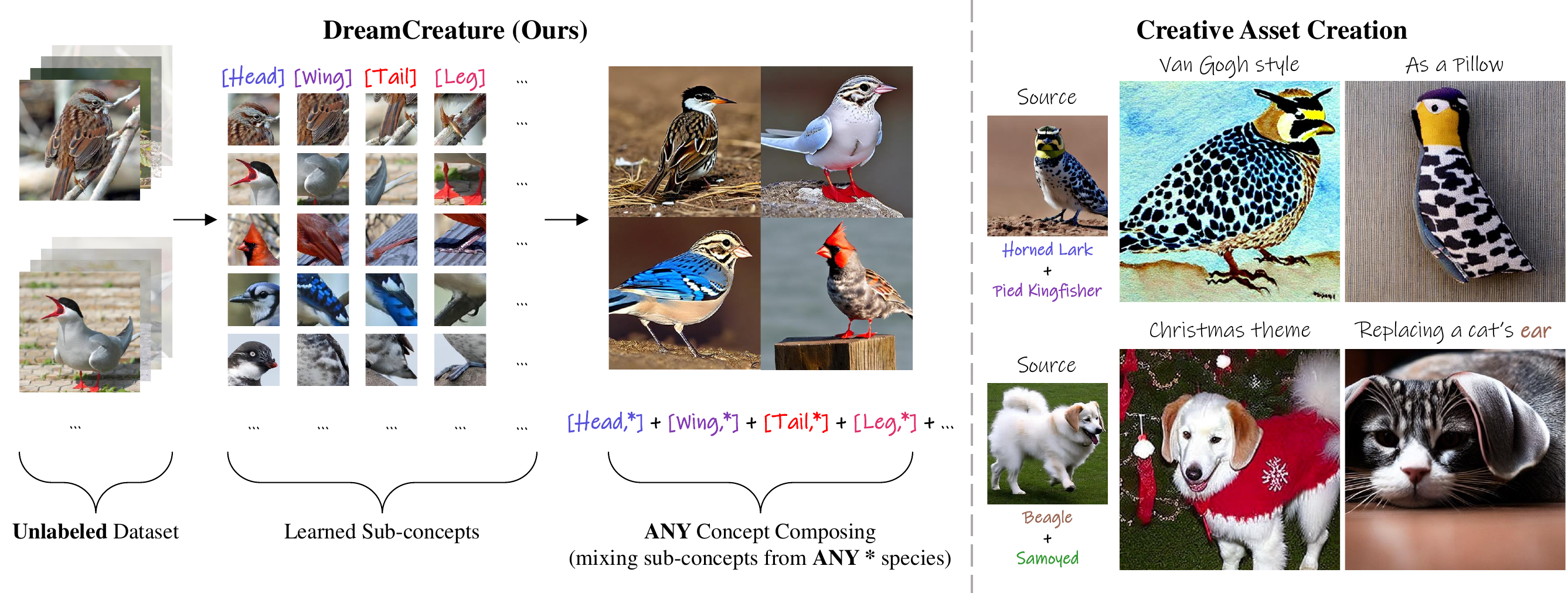}
    \captionof{figure}{\textbf{(Left)}
    Our DreamCreature generates brandy new concepts (\eg, new bird species) with photorealistic appearance and fine-grained faithful structures by seamlessly composing {\em auto-discovered} localized sub-concepts (e.g., \textit{head}, \textit{wing}) of existing concepts. \textbf{(Right)} Furthermore, these sub-concepts are utilized in creative asset generation, such as inspiring unique product designs (\eg, a themed pillow) and enabling the transfer of sub-concepts for property modifications  (\eg, replacing a cat's ear).
    }
    \label{fig:big-teaser}
\end{center}%
}]

\begin{abstract}

Recent text-to-image (T2I) generative models allow for high-quality synthesis following either text instructions or visual examples. Despite their capabilities, these models face limitations in creating new, detailed creatures within specific categories (\eg, virtual dog or bird species), which are valuable in digital asset creation and biodiversity analysis.
To bridge this gap, we introduce a novel task, {\bf Virtual Creatures Generation}: Given a set of unlabeled images of the target concepts (\eg, 200 bird species), we aim to train a T2I model capable of creating new, hybrid concepts within diverse backgrounds and contexts.
We propose a new method called {\bf DreamCreature}, which identifies and extracts the underlying sub-concepts (\eg, body parts of a specific species) in an unsupervised manner. The T2I thus adapts to generate novel concepts (\eg, new bird species) with faithful structures and photorealistic appearance by seamlessly and flexibly composing learned sub-concepts.
To enhance sub-concept fidelity and disentanglement, we extend the textual inversion technique by incorporating an additional projector and tailored attention loss regularization.
Extensive experiments on two fine-grained image benchmarks demonstrate the superiority of DreamCreature over prior methods in both qualitative and quantitative evaluation. 
Ultimately, the learned sub-concepts facilitate diverse creative applications, including innovative consumer product designs and nuanced property modifications.

\end{abstract}    
\section{Introduction}
\label{sec:intro}

\begin{figure*}[t]
    \centering
    \includegraphics[width=\textwidth]{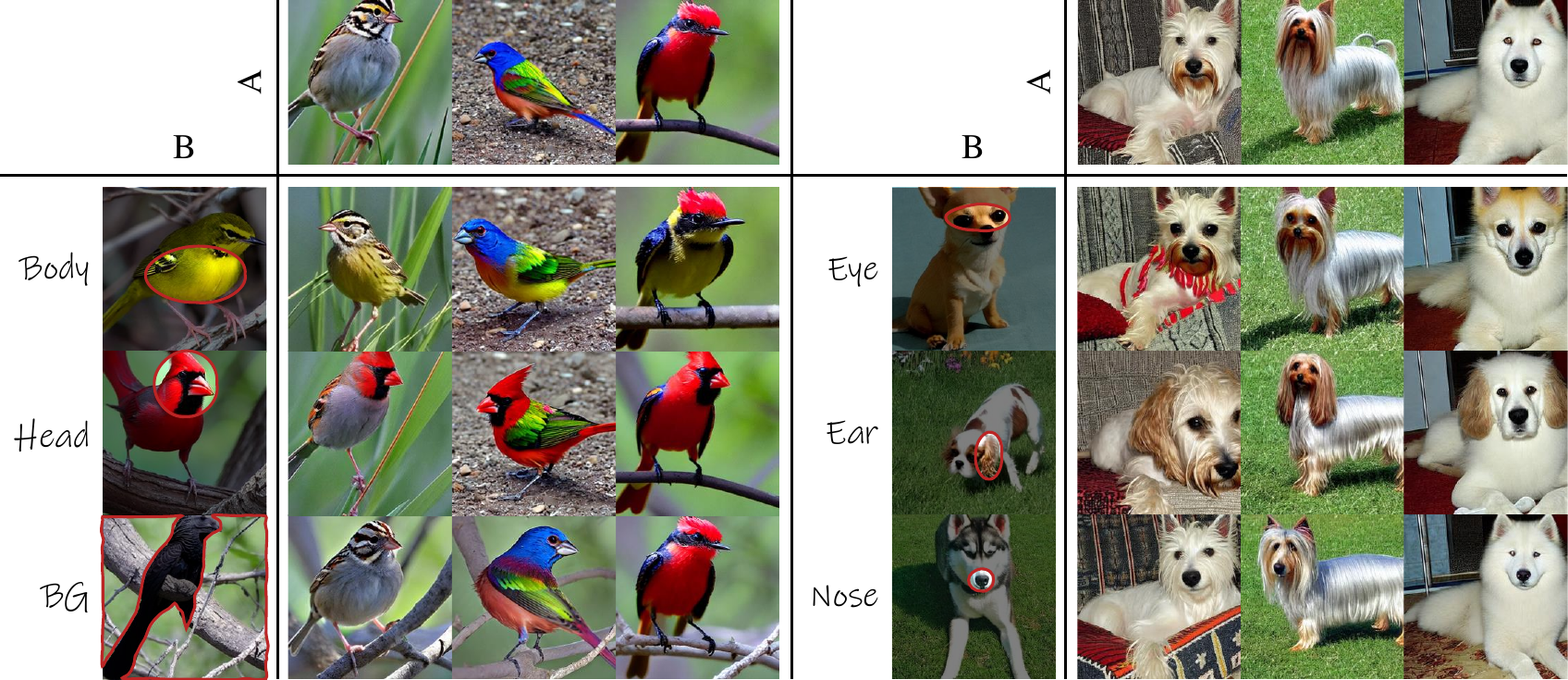}
    \caption{
    Integrating a specific sub-concept (e.g., body, head, or even background (BG) of a source concept \texttt{B} to the target concept \texttt{A}. 
    }
    \label{fig:teaser}
    \vspace{-0.4cm}
\end{figure*}

Humans possess the remarkable ability to deconstruct intricate concepts into their fundamental components and then ingeniously recombine these elements to generate novel concepts.
For instance, when presented with images of two distinct bird species, say \texttt{artic tern} and \texttt{song sparrow}, we can vividly imagine the appearance of a hybrid species with \texttt{artic tern}'s head and \texttt{song sparrow}'s body (Fig. \ref{fig:big-teaser} and \ref{fig:teaser}). This capability is likely to find potential applications in digital asset creation and biodiversity analysis \cite{vila2007origindogbreed, hebert2004identification}. However, existing generative models \cite{rombach2022ldm_sd, ramesh2022hierarchical_dalle, balaji2022ediffi, ding2022cogview2, nichol2021glide, saharia2022photorealistic_imagen} are still too limited to conduct such fine-grained synthesis tasks. 

Recent \textit{personalization} methods are applicable for this type of imagination task when they are adapted and integrated properly \cite{ruiz2023dreambooth, gal2022textualinversion, wei2023elite, kumari2023multiconcept, avrahami2023breakascene}.
Given a small set of selected images that depict the same concept, they learn a dedicated token in the textual representation space (\ie, textual inversion \cite{gal2022textualinversion}), which could be flexibly used as a proxy in various text descriptions to generate consistent images.
They focus on learning the concepts as a whole and putting them in diverse backgrounds and contexts.
This ability has been further extended to a single shot setting where only one image of the target concept is provided for learning \cite{avrahami2023breakascene}.
However, we observe a limitation in these methods regarding the reuse of their learned concepts for new concept generation at the fine-grained level (Fig.~\ref{fig:composition_compare}). 

In this work, we introduce a novel unsupervised generation task - {\em virtual creatures generation}. Given a set of unlabeled images from the target concepts (\eg, 200 bird species), we aim to train a text-to-image (T2I) generative model that can create new hybrid concepts in diverse backgrounds and contexts. 
To realize this task, we further formulate a {\bf DreamCreature} method.
It is capable of leveraging off-the-shelf image encoders (\eg, DINO \cite{caron2021emergingdino}) to identify the underlying 
sub-concepts (\eg, body parts of a specific bird species such as wings, head, tail). It further adapts the T2I model to generate these sub-concepts through textual inversion.
To further improve sub-concept fidelity and disentanglement, we introduce a projector for mapping sub-concept tokens to the text embedding space, complemented by tailored attention loss regularization. This attention loss serves a dual purpose: it not only ensures accurate positioning of sub-concepts in the cross-attention map but also enforces that each image region is occupied by no more than one sub-concept.

Our contributions are as follows:
{\bf (i)} We introduce a more challenging fine-grained image generation task -- \textit{virtual creatures generation}, that needs to create new hybrid concepts based on learned sub-concepts. 
This not only reveals the limitations of existing generative models but also expands the scope of generative AI. 
{\bf (ii)} 
We propose a novel method called \textbf{DreamCreature}, capable of automatically discovering the underlying 
sub-concepts in an unsupervised manner and allowing flexible generation of new concepts with faithful holistic structures and photorealistic appearance. 
{\bf (iii)} 
To benchmark this generation task, we introduce two quantitative metrics. Extensive experiments on CUB-200-2011 (birds) and Stanford Dogs datasets show that DreamCreature excels over prior art alternatives in both qualitative and quantitative assessments.
{\bf (iv)} 
Finally, DreamCreature opens up new avenues for creative applications such as innovative consumer product design and nuanced property modifications, showcasing the practical utility and versatility of the learned sub-concepts.
 
\section{Related Work}

\begin{figure*}[t]
    \centering
    \includegraphics[width=\textwidth]{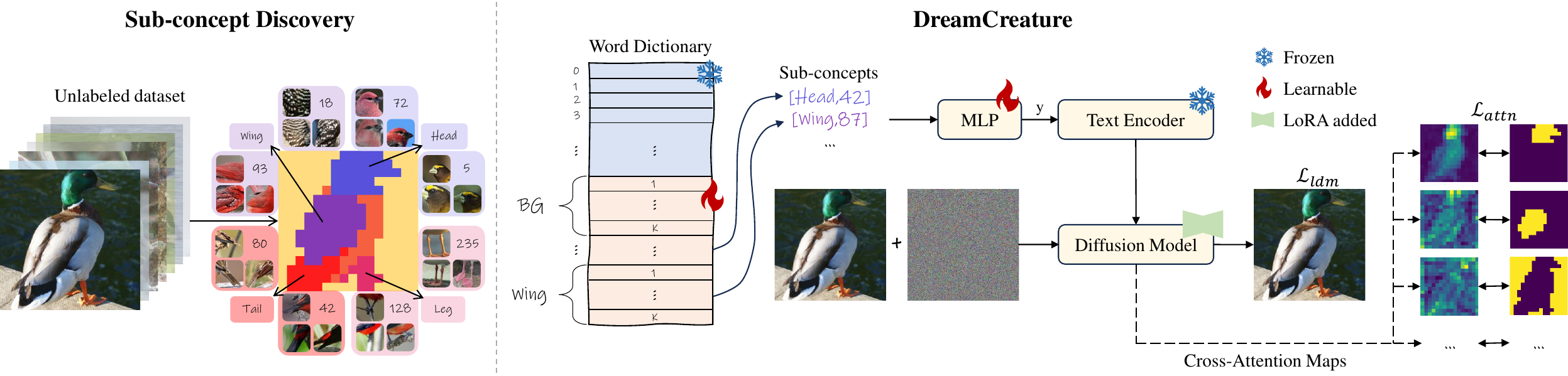}
    \caption{
    Overview of our DreamCreature. 
    \textbf{(Left)} 
    Discovering sub-concepts within a semantic hierarchy involves partitioning each image into distinct parts and forming semantic clusters across unlabeled training data. 
    \textbf{(Right)} 
    These clusters are organized into a dictionary, and their semantic embeddings are learned through a textual inversion approach. For instance, a text description like ``a photo of a [Head,42] [Wing,87]...'' guides the optimization of the corresponding textual embedding by reconstructing the associated image. To promote disentanglement among learned concepts, we minimize a specially designed attention loss, denoted as $\mathcal{L}_{attn}$.
    }
    \label{fig:coco-arch}
    \vspace{-0.4cm}
\end{figure*}

\paragraph{Text-to-image synthesis and personalization.} 
State-of-the-art large text-to-image (T2I) diffusion models \cite{rombach2022ldm_sd, ramesh2022hierarchical_dalle, balaji2022ediffi, ding2022cogview2, nichol2021glide, saharia2022photorealistic_imagen, ding2021cogview} have surpassed conventional methods \cite{zhu2019dmgan, tao2022dfgan, xu2018attngan, yin2019semantics, qiao2019mirrorgan, li2019controllable} in generating high-quality images from unstructured text prompts. These advanced generative models have been widely applied to global \cite{brooks2022instructpix2pix, mokady2023nulltext, kawar2023imagic} and localized \cite{couairon2023diffedit, yang2023paint, hertz2022prompttoprompt, avrahami2023blended} image editing tasks.

However, the effectiveness of T2I models is constrained by the user's ability to articulate their desired image through text. These models face challenges in faithfully replicating visual characteristics from a reference set and generating innovative interpretations in diverse contexts, even with detailed textual descriptions.

To address this limitation, various \textit{personalization} techniques have been developed. These techniques obtain a new single-word embedding from multiple images \cite{gal2022textualinversion, voynov2023pplus, alaluf2023neti, ruiz2023dreambooth} or multiple new word embeddings for various subjects within a single image \cite{avrahami2023breakascene} through inversion. However, most of these approaches focus on extracting the complete subjects as presented but not creating new subjects.
We address this limitation in this work by introducing an unsupervised task
of creating novel concepts/subjects from existing ones 
by meaningful recombination (see Fig.~\ref{fig:teaser}).

\vspace{-0.2cm}

\paragraph{Creative editing and generation.} Creativity involves generating innovative ideas or artifacts across various domains \cite{cetinic2022understanding}. Extensive research has explored the integration of creativity into Generative Adversarial Networks (GANs) \cite{ahmed2017can, nobari2021creativegan, sbai2019design} and Variational Autoencoders (VAEs) \cite{das2020creativedecoder, cintas2022creativity}. For example, DoodlerGAN \cite{ge2020doodlergan} learns and combines fine-level part components to create sketches of new species.

A recent study by \cite{vinker2023conceptdecomposition} demonstrated decomposing personalized concepts into distinct visual aspects, creatively recombined through diffusion models. InstructPix2Pix \cite{brooks2022instructpix2pix} allows creative image editing through instructions, while ConceptLab \cite{richardson2023conceptlab} aims to identify novel concepts within a specified category, deviating from existing concepts.

In contrast, our focus is on training a text-to-image generative model that creatively generates new concepts by seamlessly composing sub-concepts from different existing concepts in diverse backgrounds.

\section{Methodology}

\paragraph{Overview.}
Given an unlabeled image dataset for target concepts
(\eg, 200 bird species \cite{wah2011cub200}), we aim to train a text-to-image generative model that can create new hybrid concepts 
in diverse backgrounds and contexts.
To that end, we propose a novel {\bf DreamCreature} method, as depicted in Fig \ref{fig:coco-arch}.

It starts by discovering the underlying 
sub-concepts (\eg, body parts of each species) in a two-tier hierarchy, as detailed in Sec.~\ref{sec:kmeans}.
Note that \textit{each target concept is composed of a set of sub-concepts}.
Paired with the training images $\{x\}$, this semantic hierarchy subsequently serves as the supervision to fine-tune a pre-trained text-to-image model, say a latent diffusion model \cite{rombach2022ldm_sd}, denoted as
$\{\epsilon_\theta, \tau_\theta, \mathcal{E}, \mathcal{D}\}$,
where $\epsilon_\theta$ represents the diffusion denoiser, $\tau_\theta$ the text encoder, and $\mathcal{E}/ \mathcal{D}$ the autoencoder respectively.
We adopt the textual inversion technique \cite{gal2022textualinversion}.
Concretely, we learn a set of pseudo-words $p^*$ for each sub-concept in the word embedding space with:
\begin{align} 
    \mathcal{L}_{ldm} &= \mathbb{E}_{z,t,p,\epsilon}\big[ || \epsilon - \epsilon_\theta(z_t,t,\tau_\theta(y_{p})||^2_2 \big], \label{eq:diffusionloss} \\
    p^* & = \argmin_p \, \mathcal{L}_{ldm} \label{eq:textinv_obtain},
\end{align}
where $\epsilon \sim \mathcal{N}(0,1)$ denotes the unscaled noise, $t$ is the time step, $z = \mathcal{E}(x)$ is the latent representation of the image, $z_t$ is the latent noise at time $t$, and $y_{p}$ is the text condition that includes $p$ as part of the text tokens. $\mathcal{L}_{ldm}$ is a standard diffusion loss \cite{ho2020ddpm} to reconstruct the sub-concepts. As each target concept is composed of a set of sub-concepts, its reconstruction is achieved by the reconstruction of the associated set of sub-concepts.

\subsection{Sub-concepts Discovery}
\label{sec:kmeans}

To minimize the labeling cost, we develop a scalable process
to reveal the underlying semantic hierarchy with sub-concepts in an unsupervised fashion.
We leverage the off-the-shelf vision model for image decomposition and clustering.
Specifically, given an image $x_i$, we employ DINO \cite{caron2021emergingdino} to extract the feature map $F=\{F_i = \mathrm{dino}(x_i)\}_i^N$.
We then conduct three-level hierarchical clustering (see Fig. \ref{fig:coco-arch}):
\begin{enumerate}
    \item At the top level, $k$-means is applied with two clusters
    on the feature maps $F$ to obtain the foregrounds and backgrounds $B$.
    \item At the middle level, $k$-means is further applied
    on the foregrounds to acquire $M$ clusters each representing a class-agnostic sub-concept, such as the head of birds.
    \item At the bottom level, we further group each of the $M$ clusters as well as the background cluster $B$ into $K$ splits. Each split refers to further fine meaning, such as the head of a specific bird species, or a specific background style.
\end{enumerate}
After the above structural analysis, each region of an image will be tagged with the corresponding cluster index.
We represent these cluster tags as follows:
\begin{align} \label{eq:concept}
    p = (0,k_0), (1, k_1), ..., (M, k_M),
\end{align}
where the first pair refers to the background style,
and the following $M$ pairs denote the combinations of 
$M$ sub-concepts (\eg, head, body, wings) each associated with a specific concept (\eg, sparrow), and $k \in \{1, \ldots, K\}$.
This description will be used as the textual prompt in model training, such as ``a photo of a [$p$]''.
Please refer to the supplementary material for more examples of the auto-discovered semantic hierarchy. This process also yields the segmentation mask of each $m$-th sub-concept, which we define as $S_m$.

\subsection{Sub-concepts Projection}
\label{sec:mapper}

In contrast to prior text inversion studies \cite{gal2022textualinversion}, our task requires mastering a greater quantity—specifically, $(M+1)K$—of word tokens derived from a collection of self-discovered concepts and sub-concepts marked by inherent imperfections (such as partial overlap and over splitting). This makes the learning task more demanding. To enhance the learning process, we propose a neural network $f$ comprising a two-layer MLP with ReLU activation:
\begin{align} \label{eq:token_embs}
    y_p = f(e(p)),
\end{align}
where $y_p$ will be subsequently used as the input\footnote{Word templates such as ``a photo of a '' will be used.} to the text encoder $\tau_\theta$ and
$e \in \mathbb{R}^{MK \times D}$ is a learnable word embedding dictionary that maps $p$ to their respective embeddings. 

Our design demonstrates quicker convergence than directly learning the final word embeddings $e(.)$ \cite{gal2022textualinversion} (see Sec.~\ref{sec:abl}). This could be attributed to the entanglement of word embeddings in the conventional design, where there is no information exchange among them during optimization. This lack of communication leads to lower data efficiency and slower learning. It's worth noting that the conventional design is a specific instance of our approach when $f$ is an identity function.

\subsection{Model Training}\label{sec:learningobjective}

Fine-tuning the T2I model, rather than sorely learning pseudo-words, has been shown to achieve better reconstruction of target concepts as demonstrated in \cite{ruiz2023dreambooth, kumari2023multiconcept}. However, this comes with a significant training cost. Thus, we apply LoRA (low-rank adaptation) \cite{hu2022lora} to the cross-attention block for efficient training. We then minimize the diffusion loss $\mathcal{L}_{ldm}$ (Eq.~\eqref{eq:diffusionloss}) to learn both pseudo-words and LoRA adapters.

While training with only $\mathcal{L}_{ldm}$, entanglement happens between parts, as evident from the attention maps in the cross-attention block of the denoiser $\epsilon_\theta$ (see Sec.~\ref{sec:abl}). This entanglement arises due to the correlation between sub-concepts (\eg, a bird head code is consistently paired with a bird body code to represent the same species). To address this issue, we introduce an entropy-based attention loss as regularization:
\begin{align} \label{eq:attn_loss}
    \mathcal{L}_{attn} &= \mathbb{E}_{z,t,m} \big[ -\big(S_{m} \log \hat{A}_m + \nonumber \\
    &\quad\quad\quad\quad\quad\,\,\,\,(1 - S_{m}) \log (1 - \hat{A}_m)\big) \big], \\
    \bar{A}_m &= \frac{1}{L}\sum_l^L A_{l,m} , \quad \hat{A}_{m,i,j} = \frac{\bar{A}_{m,i,j}}{\sum_k \bar{A}_{k,i,j}},
\end{align}
where $A \in [0,1]^{M \times HW}$ represents the cross-attention map between the $m$-th sub-concept and the noisy latent $z_t$, $L$ represents the number of selected attention maps, $\hat{A} \in [0,1]^{M \times HW}$ represents the averaged and normalized cross-attention map over all sub-concepts and $S_m \in \{0,1\}^{M \times HW}$ serves as the mask that indicates the location of $m$-th part. In cases where the sub-concept is not present in the image (\eg, occluded), we set both $S_m$ and $\hat{A}_m$ as 0 to exclude them.

\begin{figure*}[t]
    \centering
    \includegraphics[width=\linewidth]{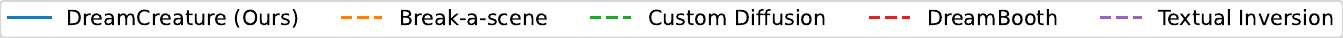} \\
    \begin{subfigure}[b]{0.49\textwidth}
         \centering
         \includegraphics[width=0.49\linewidth]{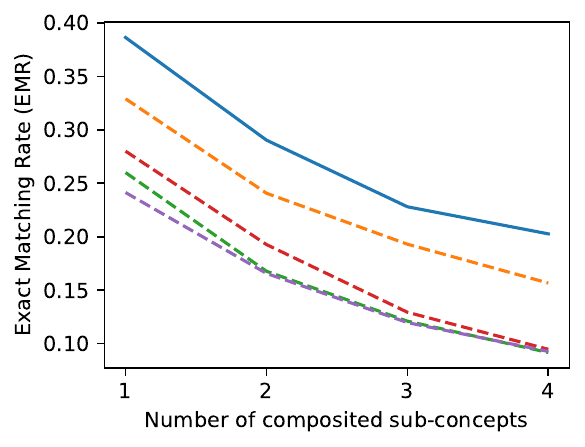}
         \hfill
         \includegraphics[width=0.49\linewidth]{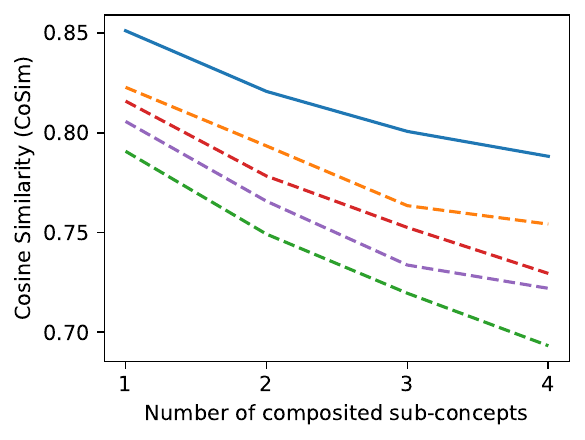}
         \caption{Bird Generation using CUB-200-2011}
         \label{fig:bird_composition}
     \end{subfigure}
     \hfill
     \begin{subfigure}[b]{0.49\textwidth}
         \centering
         \includegraphics[width=0.49\linewidth]{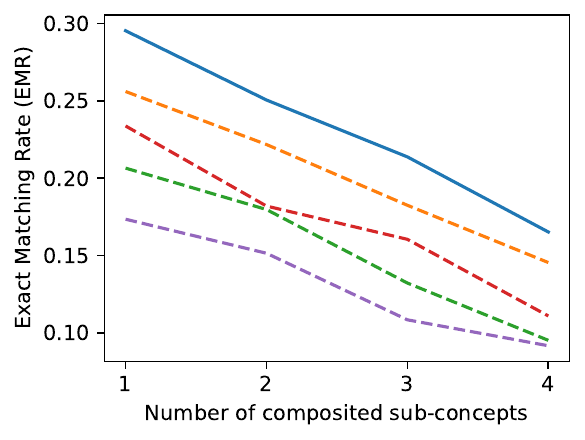}
         \hfill
         \includegraphics[width=0.49\linewidth]{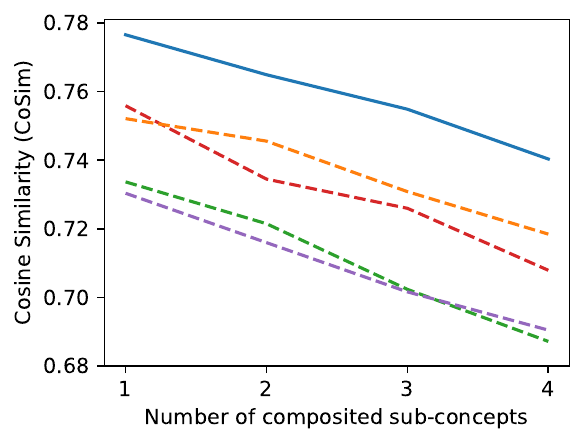}
         \caption{Dog Generation using Stanford Dogs}
         \label{fig:dog_composition}
     \end{subfigure}
    \caption{Quantitative comparisons of virtual creature generation in terms of EMR and CoSim.}
    \vspace{-0.3cm}
    \label{fig:composition_compare}
\end{figure*}

Thus, the overall learning objective is defined as:
\begin{align}
    \mathcal{L}_{total} = \mathcal{L}_{ldm} + \lambda_{attn} \mathcal{L}_{attn},
\end{align}
where $\lambda_{attn}=0.01$. 

We focus on attention maps at the resolution of $16 \times 16$ where rich semantic information is captured \cite{hertz2022prompttoprompt}. Normalization is performed at each location to ensure that the sum of a patch location equals 1. This attention loss aims to maximize the attention of a specific sub-concept at a particular location which implicitly minimizing the attention of other sub-concepts. Compared to the mean-square based attention loss \cite{avrahami2023breakascene}, this intuitively ensures that a sub-concept only appears once at a particular location, facilitating stronger disentanglement from other sub-concepts during the denoising operation. When generating a sub-concept for a particular location, the diffusion model $\epsilon_\theta$ should only attend to the sub-concept instead of other non-related sub-concepts.
We have validated the effectiveness of this design empirically (see Fig. \ref{fig:abl_comparison}). 

\section{Experiments}

\paragraph{Datasets.} We demonstrate virtual creature generation on two fine-grained object datasets: CUB-200-2011 (birds) \cite{wah2011cub200} which contains 5994 training images, and Stanford Dogs \cite{khosla2011dog} which contains 12000 training images.

\begin{table}[t]
    \centering
    \adjustbox{max width=\linewidth}{
        \begin{tabular}{c|cccc}
             & Learnable & \multirow{2}{*}{Fine Tune} & \multirow{2}{*}{Disentanglement} & \multirow{2}{*}{Projector} \\
             & Token & & & \\
             \midrule
            Textual Inversion \cite{gal2022textualinversion} & \cmark & \xmark & \xmark & \xmark \\
            DreamBooth \cite{ruiz2023dreambooth}  & \cmark & LoRA* & \xmark & \xmark \\
            CustomDiffusion \cite{kumari2023multiconcept}  & \cmark & $K/V$ & \xmark & \xmark \\
            Break-a-scene \cite{avrahami2023breakascene}  & \cmark & LoRA*  & MSE & \xmark \\
            Ours & \cmark & LoRA & Eq.~\ref{eq:attn_loss} & \cmark \\
        \end{tabular}
    }
    \caption{Comparing our and competing methods in design properties.
    *: We fine-tuned the added LoRA \cite{hu2022lora} adapter rather than the entire diffusion model $\epsilon_\theta$ due to resource limit. MSE is a mean-square based attention loss used in \cite{avrahami2023breakascene}.
    }
    \vspace{-0.2cm}
    \label{tab:summary_methods}
\end{table}

\vspace{-0.4cm}

\paragraph{Implementation.} 
For virtual creature generation, we assess the model's ability to combine up to 4 different parts from 4 distinct species/images. 
We set $M=5$ for bird generation (head, front body/breast area, wings, legs, tail) and $M=7$ for dog generation (forehead, eyes, mouth/nose, ears, neck, body/tail, legs). For both datasets, $K$ is set as 256, ensuring sufficient coverage of all fine-grained classes (\ie, 200 for birds and 120 for dogs). 
We randomly generate 500 images by sampling 500 sets of sub-concepts. For each image, we randomly replace an original sub-concept with any sub-concept from another 500 non-overlapping sets of sub-concepts. The resulting set of sub-concepts may take the form of ``$(0,k_A)$ $(1,k_B)$ $(2,k_C)$ ... $(M,k_D)$'', representing a composition from species A, B, C, and D. 
Stable Diffusion v1.5 \cite{rombach2022ldm_sd} is used. Please see the supplementary material for further training details.

\vspace{-0.4cm}

\paragraph{Competitors.} We compare our method with the recent personalization methods: Textual Inversion \cite{gal2022textualinversion}, DreamBooth \cite{ruiz2023dreambooth}, Custom Diffusion \cite{kumari2023multiconcept}, Break-a-scene \cite{avrahami2023breakascene}. 
These personalization methods were designed to take single or multiple images with associated labeled concepts as input. 
For fair comparison, we adapted them to our problem setting by inputting the same unsupervised associated concepts as DreamCreature (obtained as in Sec.~\ref{sec:kmeans}). Thus, the text prompt for each image is ``a photo of [$p$]'' where $p$ is expressed in Eq.~\eqref{eq:concept}.
We employ the official implementations released by the original authors for training.
We summarize the main design properties of all competitors in Tab.~\ref{tab:summary_methods}.

\begin{figure*}[t]
    \centering
    \caption{Visual comparison on 4-species (specified on the top row) hybrid generation. The last column indicates generated images with different styles (\ie, \textit{DSLR}, \textit{Van Gogh}, \textit{Oil Painting}, \textit{Pencil Drawing}).}
    \includegraphics[height=0.92\textheight]{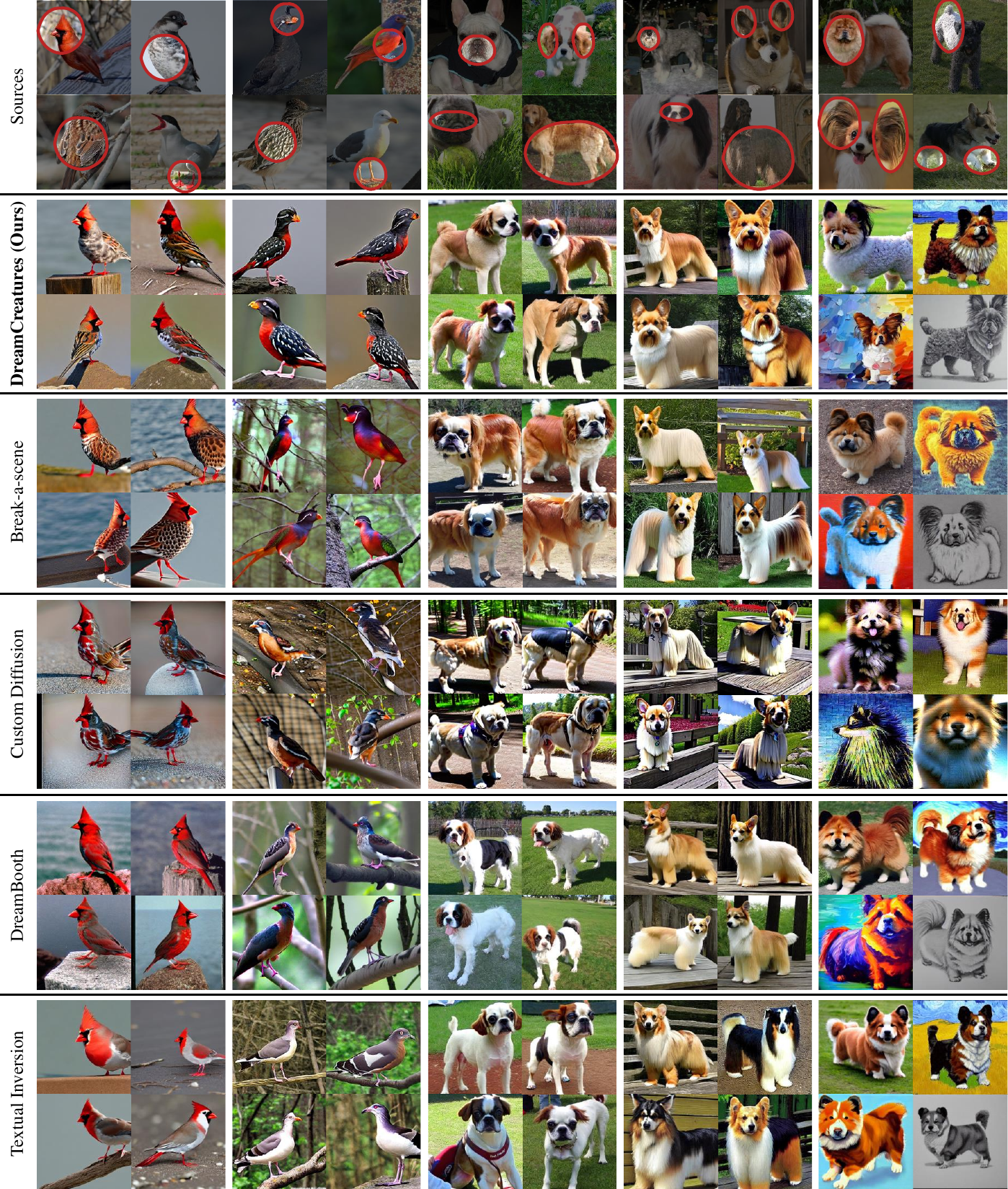}
    \label{fig:composition_visualization}
\end{figure*}

\subsection{Virtual Creature Generation Evaluation} \label{sec:vcceval}

\textbf{Evaluation metrics.} To assess a model's ability to disentangle and composite sub-concepts, we introduce two metrics: (a) exact matching rate (\textit{EMR}) and (b) cosine similarity (\textit{CoSim}) between the $k$-means embeddings of the sub-concepts of real and generated images. Utilizing the pre-trained $k$-means from Sec.~\ref{sec:kmeans}, we predict the sub-concepts of generated images. \textit{EMR} quantifies how accurately the cluster index of sub-concepts of generated images matches the sub-concepts of the corresponding real images whereas \textit{CoSim} measures the cosine similarity between the $k$-means centroid vector that the sub-concept belongs to between generated and real images.
These metrics assess the model's ability to follow the input sub-concepts and accurately reconstruct them, with perfect disentanglement indicated by \textit{EMR} of 1 and \textit{CoSim} of 1. A detailed algorithm is provided in the supplementary material.

\vspace{-0.4cm}

\begin{figure*}[t]
    \centering    
    \begin{subfigure}[b]{0.196\linewidth}
        \includegraphics[width=\linewidth]{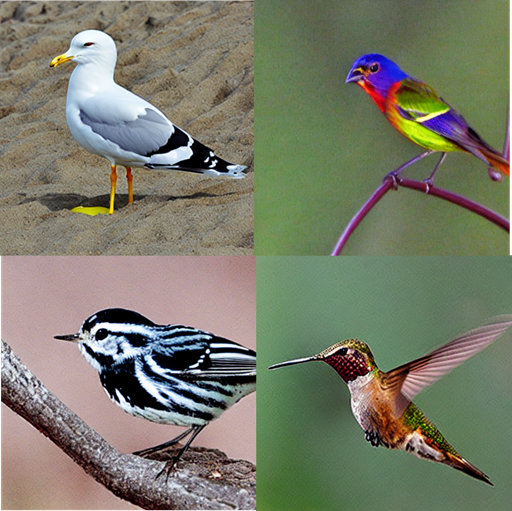} \\
        \includegraphics[width=\linewidth]{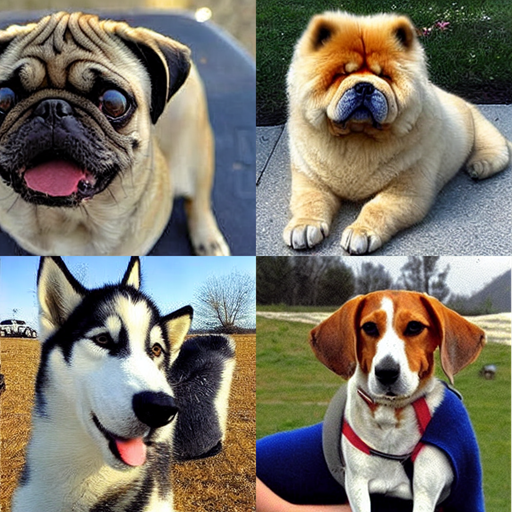}
        \caption{Textual Inversion}
    \end{subfigure}
    \begin{subfigure}[b]{0.196\linewidth}
        \includegraphics[width=\linewidth]{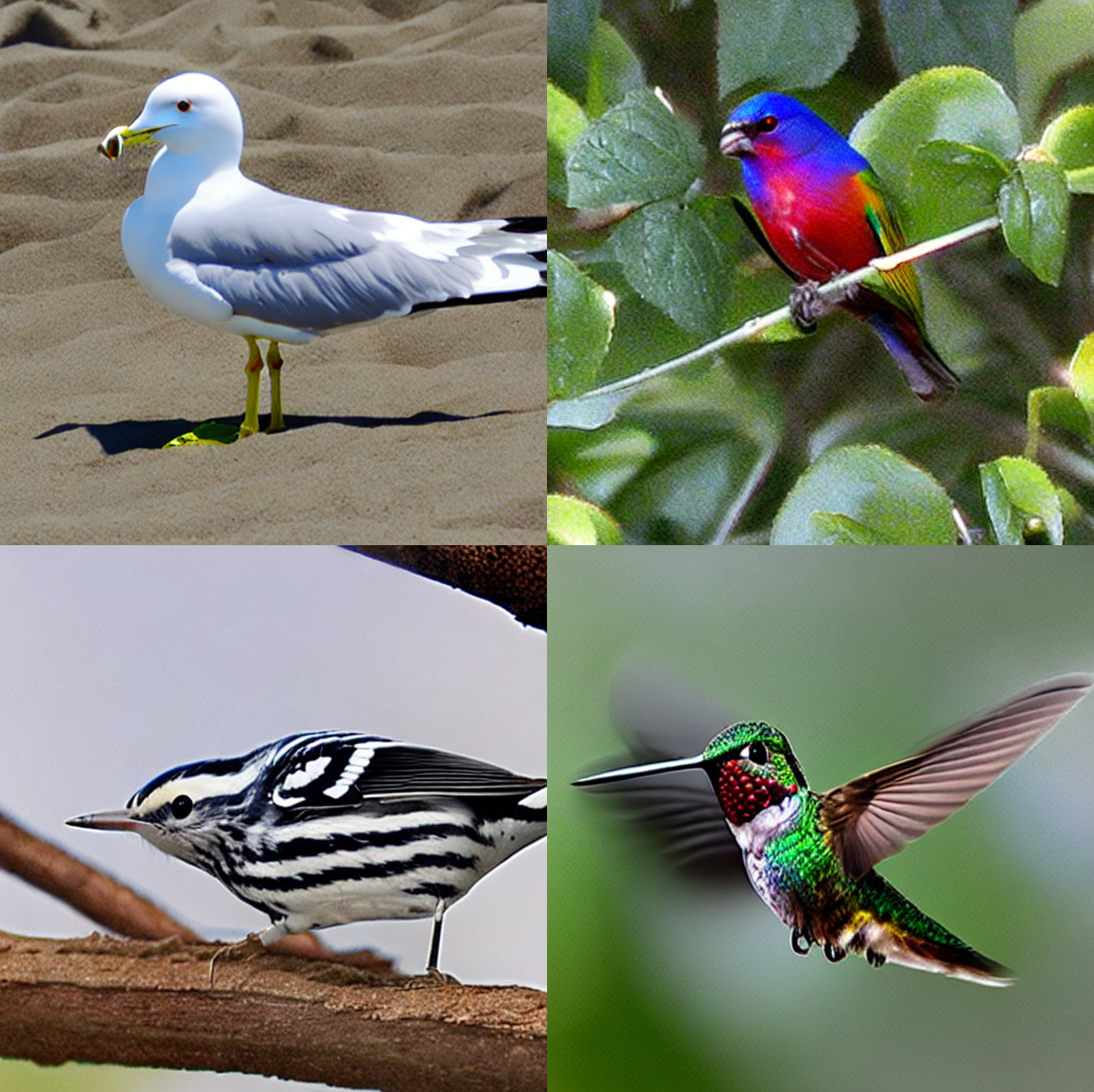} \\   
        \includegraphics[width=\linewidth]{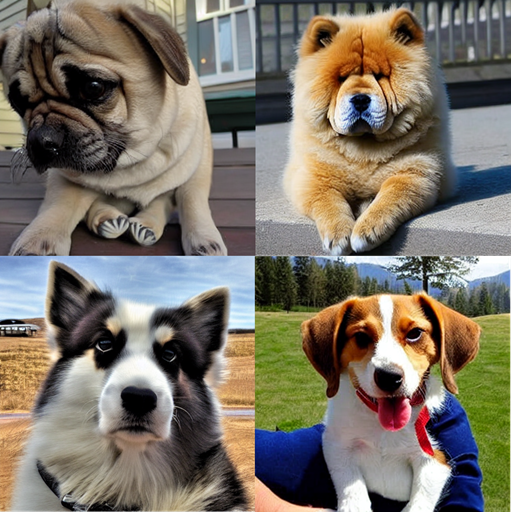}
        \caption{DreamBooth}
    \end{subfigure}
    \begin{subfigure}[b]{0.196\linewidth}
        \includegraphics[width=\linewidth]{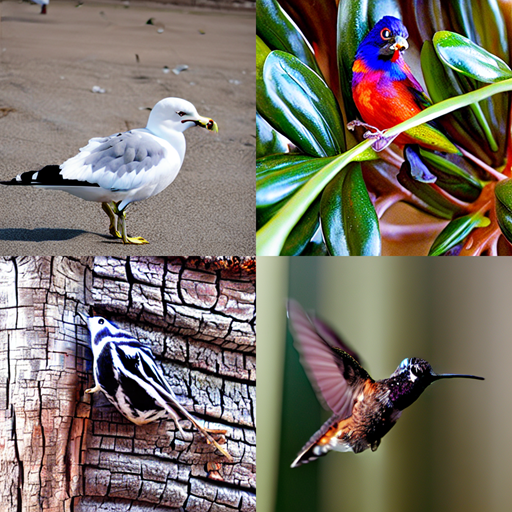} \\   
        \includegraphics[width=\linewidth]{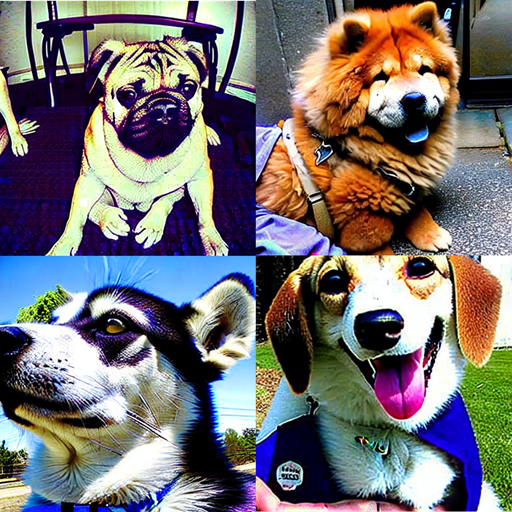}
        \caption{Custom Diffusion}
    \end{subfigure}
    \begin{subfigure}[b]{0.196\linewidth}
        \includegraphics[width=\linewidth]{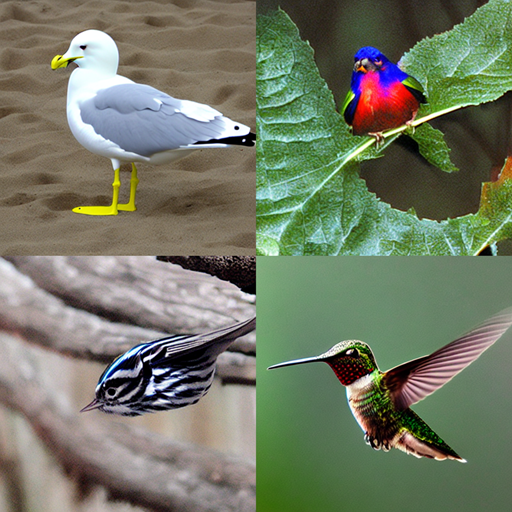} \\   
        \includegraphics[width=\linewidth]{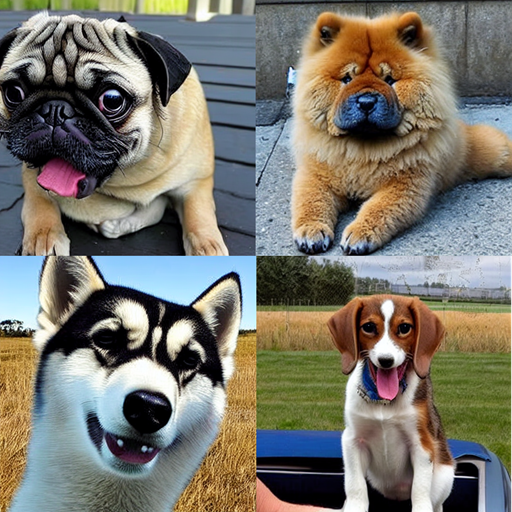}
        \caption{Break-a-scene}
    \end{subfigure}
    \begin{subfigure}[b]{0.196\linewidth}
        \includegraphics[width=\linewidth]{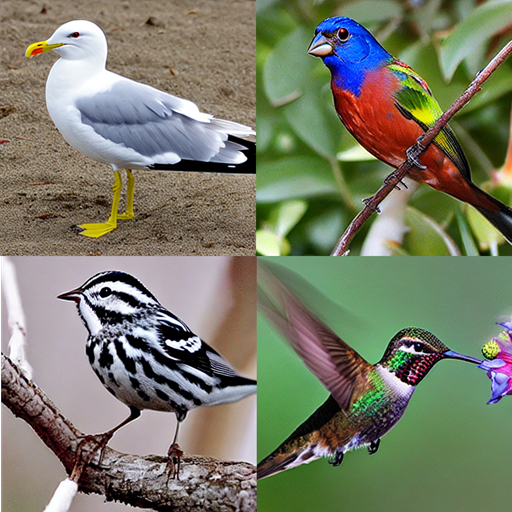} \\ 
        \includegraphics[width=\linewidth]{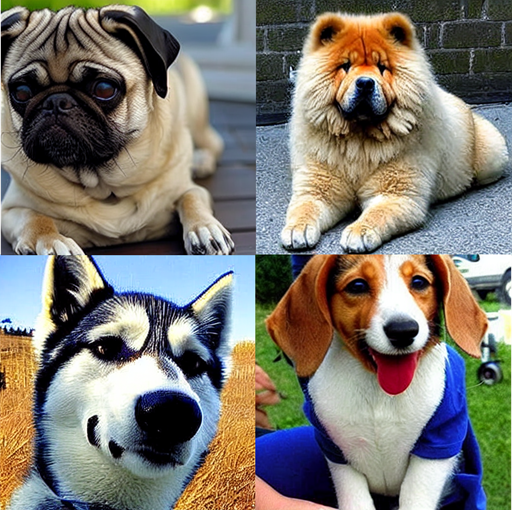}
        \caption{DreamCreature (Ours)}
    \end{subfigure}
    \caption{Visual comparison under the conventional image generation setting.
    }
    \vspace{-0.4cm}
    \label{fig:class_specific_generated_image}
\end{figure*}

\paragraph{Quantitative results.} Our findings, as shown in Fig.~\ref{fig:composition_compare}, can be summarized as follows:
\textbf{(i)} As the number of composited sub-concepts increases, EMR and CoSim decrease, reflecting the challenge of composing multiple diverse sub-concepts.
\textbf{(ii)} Break-a-scene and DreamCreature achieve notably higher EMR and CoSim scores, thanks to disentanglement through attention loss minimization.
\textbf{(iii)} DreamCreature outperforms Break-a-scene significantly by dedicated token projector and tailored attention loss designs (especially in the case of bird generation).

\vspace{-0.4cm}

\paragraph{Qualitative results.}
In Fig.~\ref{fig:composition_visualization}, we visualize the results of composing 4 different sub-concepts. While all images appear realistic, most methods struggle to assemble all 4 sub-concepts. In contrast, our methods successfully combine 4 different sub-concepts from 4 different species, demonstrating the superior ability of our approach to sub-concept composition. 
We also visualize additional examples of our method in the supplementary material. 

Furthermore, we explore the versatility of the adapted model by generating images with simple styles such as \textit{pencil drawing}. While most methods successfully incorporate specific styles into the generated image, Custom Diffusion often fails to do so, possibly due to the unconstrained fine-tuning of the cross-attention components $K/V$.

\subsection{Conventional Generation Evaluation} \label{sec:conveval}
Our method works for traditional image generation, \ie, reconstructing target concepts (\eg, specific bird species). 

\vspace{-0.4cm}

\paragraph{Evaluation metrics.} 
We measure generated image quality using FID \cite{heusel2017gans_fid} to assess model performance in terms of image distribution. 
We also compute the average pairwise cosine similarity between CLIP \cite{radford2021clip}/DINO \cite{caron2021emergingdino} embeddings of generated and real class-specific images following \cite{ruiz2023dreambooth}. Each generated image is conditioned on the sub-concepts of the corresponding real image. This results in 5,994 generated images for birds and 12,000 generated images for dogs.

\vspace{-0.4cm}

\paragraph{Quantitative results.}
In Tab.~\ref{tab:bird_score} and Tab.~\ref{tab:dog_score}, we summarize the performance of respective methods on bird and dog generation, respectively. We highlight four observations:
\textbf{(i)} Textual Inversion performs quite well compared to DreamBooth, CustomDiffusion, and Break-a-scene in terms of FID, CLIP, and DINO scores although did not fine-tune the diffusion model $\epsilon_\theta$. This may be due to the potential risk of overfitting when fine-tuning $\epsilon_\theta$ especially when learning a vast array of new concepts with many update iterations. 
It is also not uncommon to carefully tune the learning rate and the training iterations in these models when fine-tuning new concepts (\eg, only 800-1000 steps of updates to learn a new concept in \cite{avrahami2023breakascene}).
\begin{table}[t]
    \centering
    \adjustbox{max width=\linewidth}{
        \begin{tabular}{l|ccccc}
            \multirow{2}{*}{Method}& \multicolumn{5}{c}{Birds: CUB-200-2011} \\
            \cmidrule{2-6}
            & FID & CLIP & DINO & EMR & CoSim \\
            \midrule
            Textual Inversion \cite{gal2022textualinversion}  & \bf 10.10 & \bf 0.784 & 0.607 & 0.305 & 0.842 \\
            DreamBooth  \cite{ruiz2023dreambooth} & 12.94 & 0.775 & 0.594 & 0.355 & 0.856 \\
            Custom Diffusion \cite{kumari2023multiconcept} & 37.61 & 0.694 & 0.504 & 0.338 & 0.833 \\
            Break-a-Scene \cite{avrahami2023breakascene} & 20.05 & 0.742 & 0.549 & 0.390 & 0.854 \\
            \midrule
            DreamCreature (Ours) & 12.86 & 0.783 & \bf 0.618 & \bf 0.460 & \bf 0.882 \\
        \end{tabular}
    }
    \caption{Quantitative comparison for conventional generation}
    \vspace{-0.5cm}
    \label{tab:bird_score}
\end{table}
\begin{table}[t]
    \centering
    \adjustbox{max width=\linewidth}{
        \begin{tabular}{l|ccccc}
            \multirow{2}{*}{Method}& \multicolumn{5}{c}{Dogs: Stanford Dogs} \\
            \cmidrule{2-6}
            & FID & CLIP & DINO & EMR & CoSim \\
            \midrule
            Textual Inversion \cite{gal2022textualinversion} & 23.36 & 0.652 & 0.532 & 0.218 & 0.754 \\
            DreamBooth \cite{ruiz2023dreambooth} & 22.65 & 0.660 & 0.563 & 0.275 & 0.777 \\
            Custom Diffusion \cite{kumari2023multiconcept} & 42.41 & 0.593 & 0.491 & 0.253 & 0.755 \\
            Break-a-Scene \cite{avrahami2023breakascene} & 24.20 & 0.633 & 0.532 & 0.300 & 0.775 \\
            \midrule
            DreamCreature (Ours) & \bf 16.92 & \bf 0.669 & \bf 0.573 & \bf 0.358 & \bf 0.796 \\
        \end{tabular}
    }
    \caption{Quantitative comparison for conventional generation.}
    \vspace{-0.5cm}
    \label{tab:dog_score}
\end{table}
\textbf{(ii)} Nonetheless, fine-tuning the diffusion model $\epsilon_\theta$ can help improve the ability to follow prompts as shown by increased EMR and CoSim scores (\eg, EMR of at least 5\% in DreamBooth).
\textbf{(iii)} Break-a-scene has a better ability to reconstruct the sub-concepts as shown by EMR and CoSim, this is due to the attention loss explicitly forcing the sub-concept to focus on the respective semantic region.
\textbf{(iv)}
DreamCreature achieves the best performance in DINO, EMR, and CoSim scores (\eg, 7\% better in EMR compared to Break-a-scene). This indicates that not only does our image-generation ability perform comparably well with textual inversion, but DreamCreature is also able to disentangle the sub-concept learning so that it can follow the prompt instructions more accurately to generate the sub-concepts in a cohort. 
In Fig.~\ref{fig:class_specific_generated_image}, we present generated images from different methods, with CustomDiffusion exhibiting high-contrast images, possibly due to unconstrained fine-tuning on the cross-attention components $K/V$ and resulting in worse FID scores.

\begin{figure}[t]
    \centering
    \includegraphics[width=\linewidth]{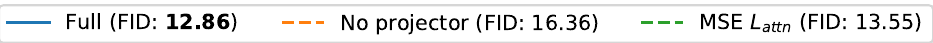} \\
     \includegraphics[width=0.49\linewidth]{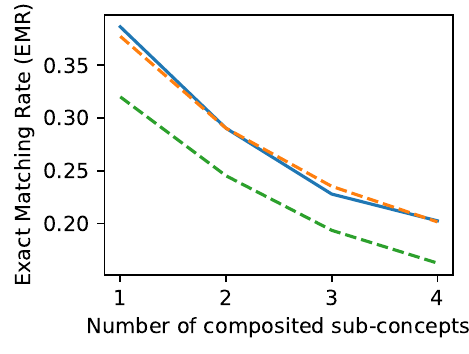}
     \hfill
     \includegraphics[width=0.49\linewidth]{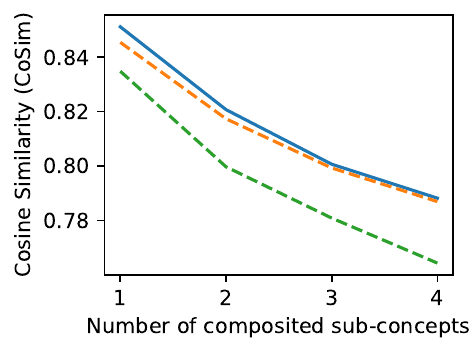}
     \caption{Ablation on our token projection and attention loss
     under the virtual creature generation setting on CUB-200-2011 birds. }
    \vspace{-0.4cm}
    \label{fig:abl_comparison}
\end{figure}
\subsection{Ablation Studies} \label{sec:abl}

\paragraph{Component analysis.} 
In Fig.~\ref{fig:abl_comparison}, 
we evaluate the effect of our proposed components (token projection and attention loss) on creating virtual bird species. \textbf{(i)} Removing the sub-concept projector outlined in Eq.~\eqref{eq:token_embs} degrades the generation quality as evidenced by a higher FID score (12.86 $\rightarrow$ 16.36) even though both EMR and CoSim remain.
\textbf{(ii)} By replacing our $\mathcal{L}_{attn}$ with the MSE loss as proposed in \cite{avrahami2023breakascene}, we observe significant deterioration in both EMR and CoSim. \textbf{(iii)} Finally, incorporating both the projector and our attention loss performs the best. This improvement highlights the necessity of incorporating interactions between multiple sub-concepts to achieve more effective sub-concept disentanglement and optimization. 

\vspace{-0.3cm}

\paragraph{Cross-attention visualization.} Our attention loss plays a crucial role in token disentanglement.
We demonstrate the impact of this loss in the Appendix \ref{sec:crossattnvis}, 
where we observe significantly enhanced disentanglement after explicitly guiding attention to focus on distinct semantic regions. 

\vspace{-0.3cm}

\paragraph{Convergence analysis.}
We present a visual comparison of images generated by various methods under the conventional setting in the Appendix \ref{sec:convergence},
spanning from the initial to the final stages of training. Notably, our DreamCreature demonstrates an ability to learn new concepts at even the early stages of training.

\begin{figure}[t]
    \centering
    \begin{subfigure}[b]{\linewidth}
        \centering
        \includegraphics[width=0.32\linewidth]{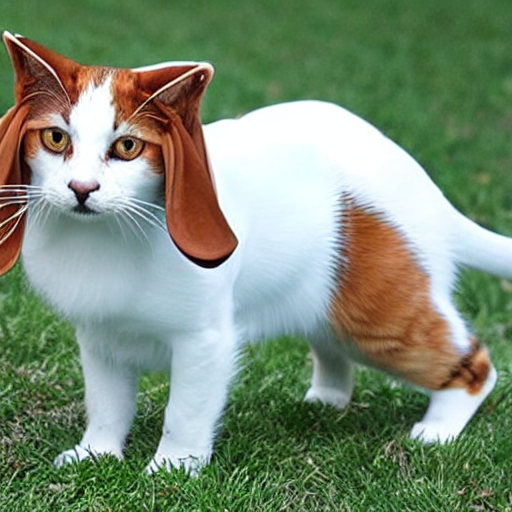}
        \includegraphics[width=0.32\linewidth]{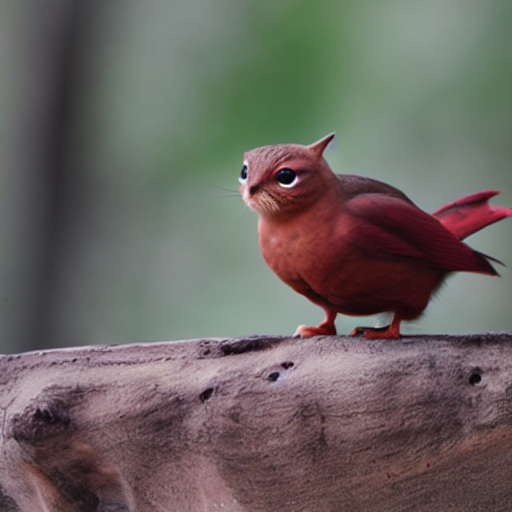}
        \includegraphics[width=0.32\linewidth]{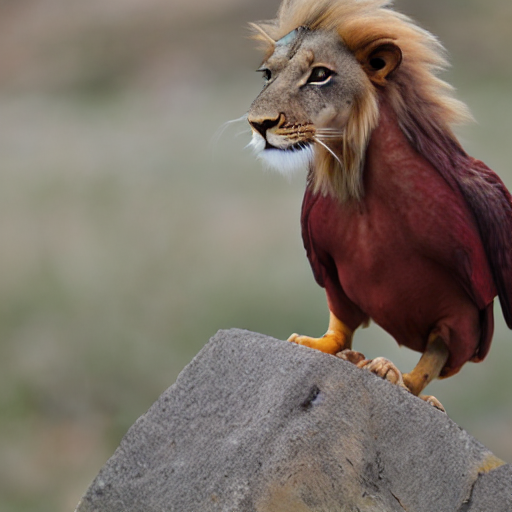}
        \caption{Sub-Concept Transfer for Property Modification} \label{fig:transfer}
    \end{subfigure}
    \begin{subfigure}[b]{\linewidth}
        \centering
        \includegraphics[width=0.32\linewidth]{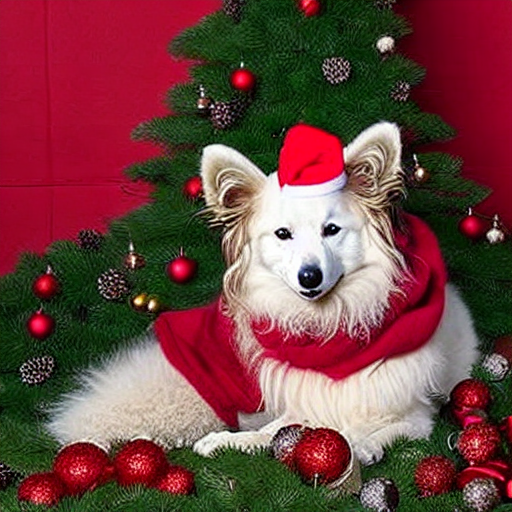}
        \includegraphics[width=0.32\linewidth]{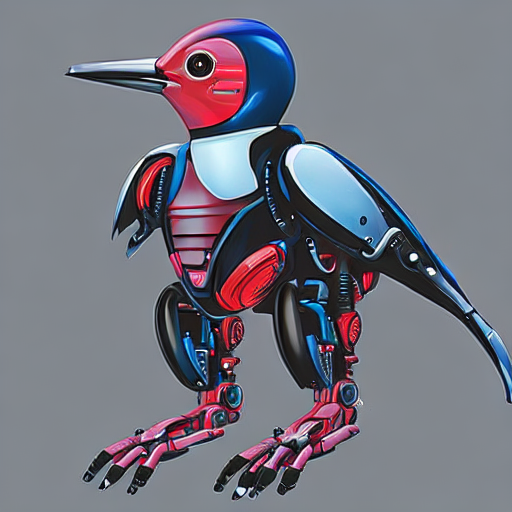}
        \includegraphics[width=0.32\linewidth]{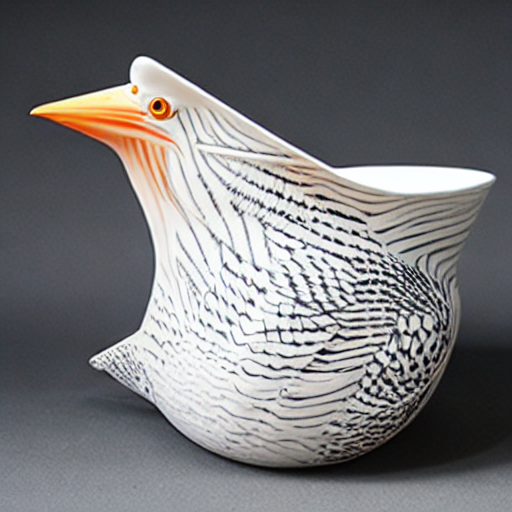}
        \caption{Creative Asset Creation} \label{fig:edit}
    \end{subfigure}
    \caption{\textbf{(a)} A cat with \texttt{beagle}'s ear; a cat with \texttt{cardinal}'s body; a lion with \texttt{cardinal}'s body. \textbf{(b)} A \texttt{samoyed} with \texttt{papillon}'s ear with Christmas theme; A robot design inspired by \texttt{red header woodpecker}'s head and \texttt{blue jay}'s body; A cup design inspired by \texttt{white pelican}'s head and \texttt{red bellied Woodpecker}'s wing texture.}
    \label{fig:application}
\end{figure}

\vspace{-0.3cm}

\paragraph{Transferability and Creative Asset Creation.} \textbf{(i)} In Fig.~\ref{fig:transfer}, we demonstrate that not only it can compose sub-concepts within the domain of the target concepts (\eg, birds), but it can also transfer the learned sub-concepts to and combine with other domains (\eg, cat). This enables the creation of unique combinations, such as a cat with a dog’s ear. \textbf{(ii)} Leveraging the prior knowledge embedded in Stable Diffusion, DreamCreature can also repurpose learned sub-concepts to design innovative digital assets. An example of this is the generation of a bird-shaped robot adorned with various sub-concepts, as depicted in Fig.~\ref{fig:edit}. These examples showcase DreamCreature's immense potential for diverse and limitless creative applications. Please see the supplementary material for more examples.

\section{Conclusion}

We introduced a novel task: Virtual Creature Generation, focusing on training a text-to-image generative model capable of seamlessly composing parts from various species to create new classes. We addressed the challenge of learning sub-concepts by proposing DreamCreature, an innovative method with token projection and tailored attention loss regularization.
DreamCreature can seamlessly compose different sub-concepts from different images, creating species that do not exist by mixing them. Extensive experiments demonstrated DreamCreature's superior performance in both qualitative and quantitative evaluation. Moreover, the learned sub-concepts demonstrate strong transferability and significant potential for creative asset generation. 
We hope that our DreamCreature will empower artists, designers, and enthusiasts to bring the creatures of their dreams to reality.

\section{Limitations and Future Works}

It is worth noting that the accuracy of obtained sub-concepts may be affected by using a self-supervised pre-trained feature extractor. Future work may explore the incorporation of encoders, such as \cite{wei2023elite}, to improve sub-concept accuracy. We also observed challenges in composing relatively small sub-concepts, like tails and legs, which require further investigation. Additionally, we are also exploring cross-domain generation, \ie, combining learned sub-concepts from different datasets to create creatures with even more diverse sub-concepts.

{
    \small
    \bibliographystyle{ieeenat_fullname}
    \bibliography{main}
}

\clearpage
\setcounter{page}{1}
\maketitlesupplementary
\appendix


\section{Implementation Details}

We conducted training on a single GeForce RTX 3090 GPU with a batch size of 2 over 100 epochs. AdamW \cite{loshchilov2018adamw} optimizer was employed with a constant learning rate of 0.0001 and weight decay of 0.01. Only random horizontal flip augmentation is used. $512 \times 512$ image resolution is applied.

We adopted the LoRA design \cite{hu2022lora} from \texttt{diffusers} library\footnote{\url{https://github.com/huggingface/diffusers/blob/main/examples/text_to_image/train_text_to_image_lora.py}}, in which the low-rank adapters were added to the $QKV$ and $out$ components of all cross-attention modules.

Regarding the attention loss (see Eq. (\textcolor{red}{5})), we selected cross-attention maps with a feature map size of $16 \times 16$. The specific layers chosen for this purpose were as follows:

{
\scriptsize
\begin{itemize}
    \item \texttt{down\_blocks.2.attentions.0.transformer\_blocks.0.attn2}
    \item \texttt{down\_blocks.2.attentions.1.transformer\_blocks.0.attn2}
    \item \texttt{up\_blocks.1.attentions.0.transformer\_blocks.0.attn2}
    \item \texttt{up\_blocks.1.attentions.1.transformer\_blocks.0.attn2}
    \item \texttt{up\_blocks.1.attentions.2.transformer\_blocks.0.attn2}
\end{itemize}
}

\section{Implementation of EMR and CoSim}

\begin{algorithm}[h]
\caption{EMR and CoSim for virtual creature generation}
\label{alg:emr_cosim_vcc}
\definecolor{codeblue}{rgb}{0.25,0.5,0.5}
\lstset{
  backgroundcolor=\color{white},
  basicstyle=\fontsize{7.2pt}{7.2pt}\ttfamily\selectfont,
  columns=fullflexible,
  breaklines=true,
  captionpos=b,
  commentstyle=\fontsize{7.2pt}{7.2pt}\color{gray},
  keywordstyle=\fontsize{7.2pt}{7.2pt}\color{codeblue},
}
\begin{lstlisting}[language=python]
# subconcept_predictor: obtained via Sec 3.1
# pipeline: diffusion generation pipeline
# real_xs: real image (Nx512x512x3) where N up to 4
# M: number of sub-concepts
# D: number of dino feature dimension

# obtain the prompt of the real image (Eq.3)
p_input = subconcept_predictor.<@\textcolor{violet}{\textbf{predict}}@>(real_xs[0])  # (M+1)
p_idxs = [0,1,...,M]
    
for real_x in real_xs[1:]:
    p_real = subconcept_predictor.<@\textcolor{violet}{\textbf{predict}}@>(real_x)  # (M+1)

    # replace one sub-concept
    rand_idx = randint(len(p_idxs))
    rand_pop = p_idxs.pop(rand_idx)
    p_input[rand_pop] = p_real[rand_pop]
    
gen_x = <@\textcolor{violet}{\textbf{pipeline}}@>(p_input)
p_gen = subconcept_predictor.<@\textcolor{violet}{\textbf{predict}}@>(gen_x)    # (M+1)

p_input_embs = subconcept_predictor.<@\textcolor{violet}{\textbf{get\_centroids}}@>(p_input)  # (M+1,D)
p_gen_embs = subconcept_predictor.<@\textcolor{violet}{\textbf{get\_centroids}}@>(p_gen)    # (M+1,D)

EMR = <@\textcolor{violet}{\textbf{average}}@>(p_input == p_gen)
CoSim = <@\textcolor{violet}{\textbf{average}}@>(<@\textcolor{violet}{\textbf{cossim}}@>(p_input_embs, p_gen_embs))
    
\end{lstlisting}
\end{algorithm}

\begin{algorithm}[h]
\caption{EMR and CoSim for conventional generation}
\label{alg:emr_cosim}
\definecolor{codeblue}{rgb}{0.25,0.5,0.5}
\lstset{
  backgroundcolor=\color{white},
  basicstyle=\fontsize{7.2pt}{7.2pt}\ttfamily\selectfont,
  columns=fullflexible,
  breaklines=true,
  captionpos=b,
  commentstyle=\fontsize{7.2pt}{7.2pt}\color{gray},
  keywordstyle=\fontsize{7.2pt}{7.2pt}\color{codeblue},
}
\begin{lstlisting}[language=python]
# subconcept_predictor: obtained via Sec 3.1
# pipeline: diffusion generation pipeline
# real_x: real image (512x512x3)
# M: number of sub-concepts
# D: number of dino feature dimension

# obtain the prompt of the real image (Eq.3)
p_real = subconcept_predictor.<@\textcolor{violet}{\textbf{predict}}@>(real_x)  # (M+1)
gen_x = <@\textcolor{violet}{\textbf{pipeline}}@>(p_real)
p_gen = subconcept_predictor.<@\textcolor{violet}{\textbf{predict}}@>(gen_x)    # (M+1)

# an example of "p" is [4, 222, 55, 23, 98, 22]
# in the "pipeline", we prepend word template like "a photo of a "
# e.g., "a photo of a [0,4] [1,222] ... [M,K]"
# the token [*,*] will be replaced by its embedding computed via Eq.4

p_real_embs = subconcept_predictor.<@\textcolor{violet}{\textbf{get\_centroids}}@>(p_real)  # (M+1,D)
p_gen_embs = subconcept_predictor.<@\textcolor{violet}{\textbf{get\_centroids}}@>(p_gen)    # (M+1,D)

EMR = <@\textcolor{violet}{\textbf{average}}@>(p_real == p_gen)
CoSim = <@\textcolor{violet}{\textbf{average}}@>(<@\textcolor{violet}{\textbf{cossim}}@>(p_real_embs, p_gen_embs))
    
\end{lstlisting}
\end{algorithm}

Our evaluation algorithms for the Exact Matching Rate (EMR) and Cosine Similarity (CoSim) between generated and real images are presented in Algorithms~\ref{alg:emr_cosim_vcc} and \ref{alg:emr_cosim}, respectively. Each algorithm is designed to evaluate a single sample. For the evaluations in Section \textcolor{red}{4.1}, 
we computed the average results over 500 iterations using Algorithm~\ref{alg:emr_cosim_vcc}. Similarly, for Section \textcolor{red}{4.2}, 
we averaged the outcomes over 5,994 and 12,000 iterations for the CUB-200-2011 (birds) and Stanford Dogs datasets, respectively.

\section{Examples of our sub-concept discovery}

\begin{figure}[h]
    \centering
    \includegraphics[width=\linewidth]{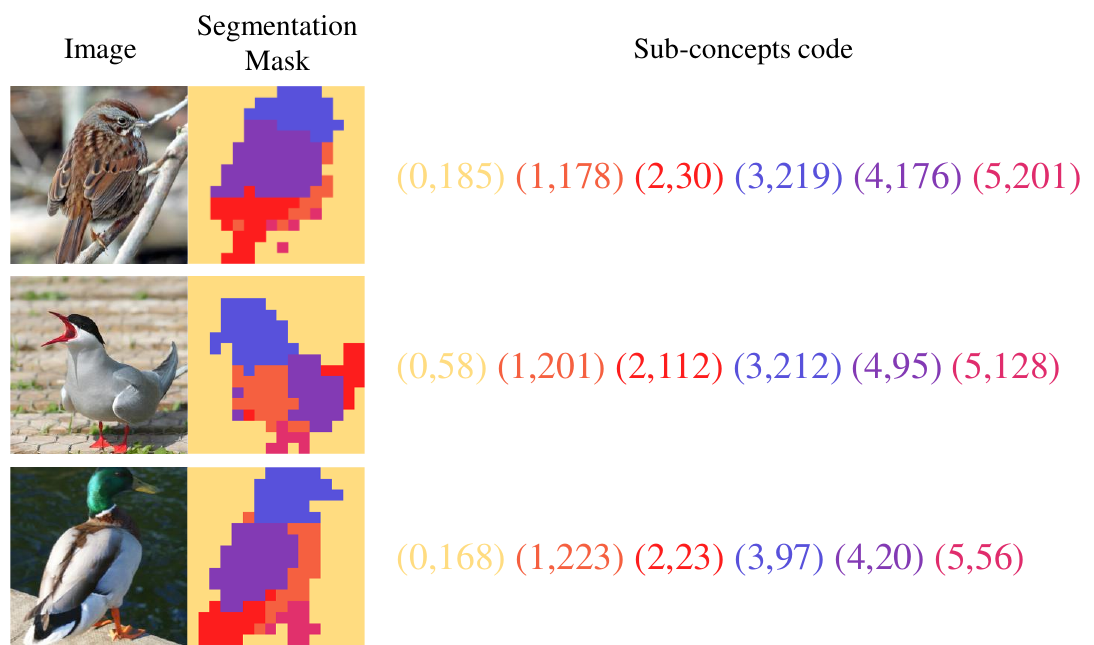}
    \caption{Three example outputs of our sub-concept discovery.}
    \label{fig:example_seg}
\end{figure}

In Fig.~\ref{fig:example_seg}, we display a few examples of our obtained segmentation masks and associated sets of sub-concepts.

\clearpage
\section{More examples}

\begin{figure}[h]
    \centering
    \begin{subfigure}[b]{\linewidth}
        \includegraphics[width=\linewidth]{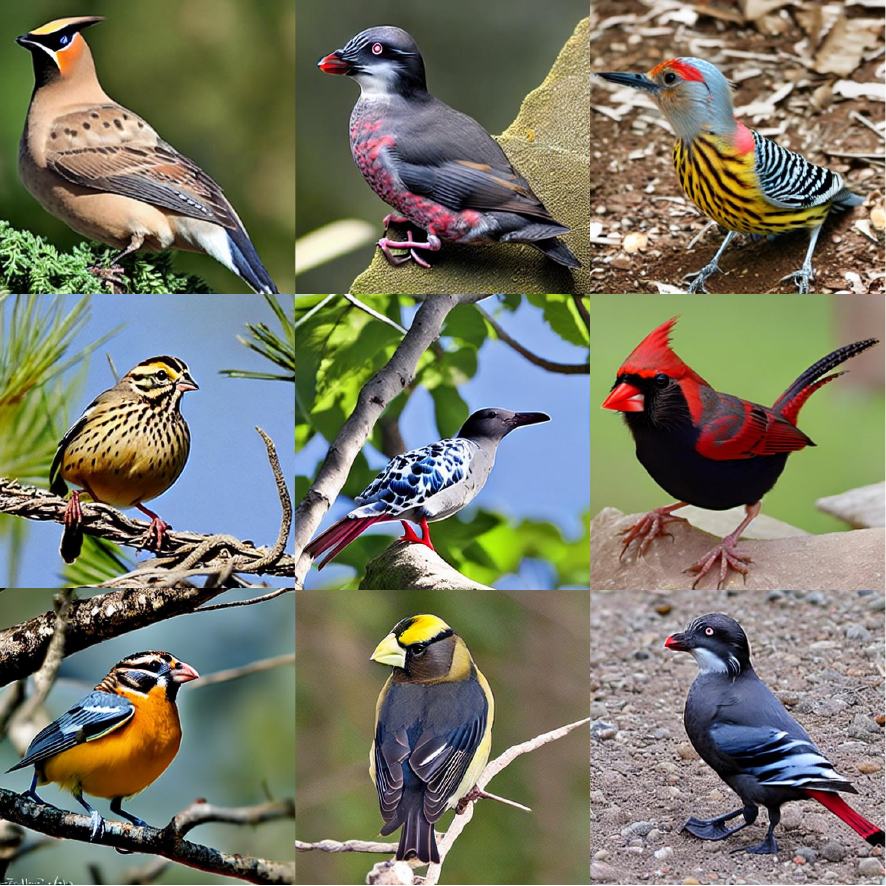}
        \caption{}
    \end{subfigure}
    \begin{subfigure}[b]{\linewidth}
        \includegraphics[width=\linewidth]{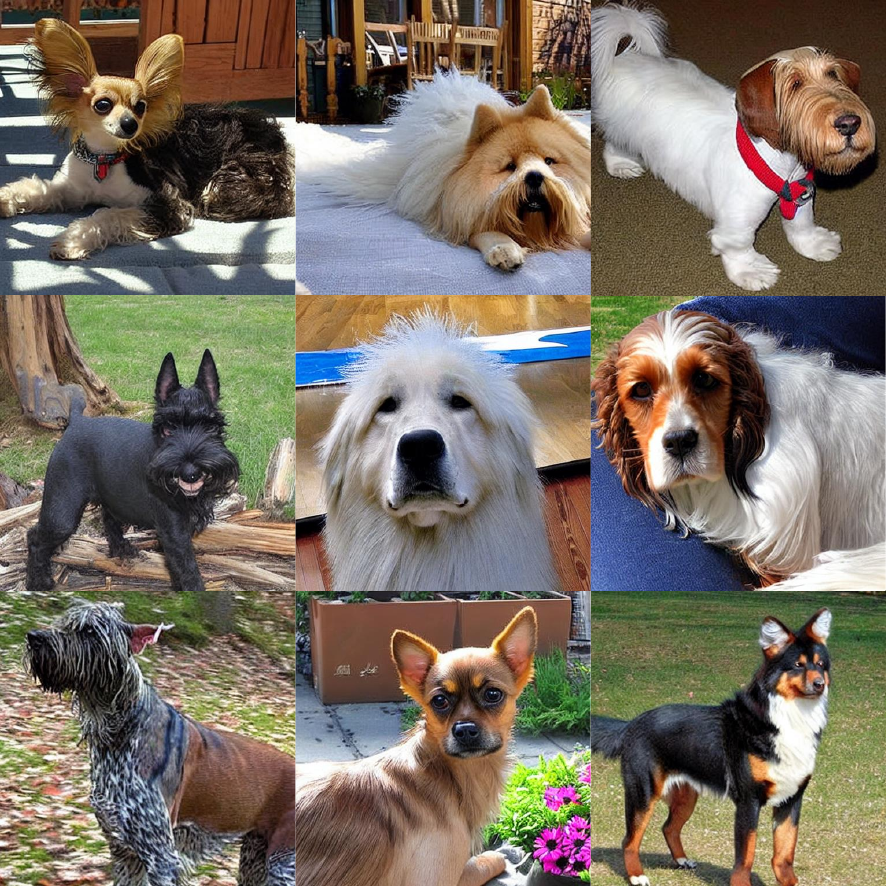}
        \caption{}
    \end{subfigure}

    \caption{We present additional examples of images generated by our DreamCreature, featuring a random selection of sub-concepts.}
    \label{fig:uncurated_bird}
\end{figure}



\begin{figure}[h]
    \centering
    \begin{subfigure}[b]{\linewidth}
        \includegraphics[width=\linewidth]{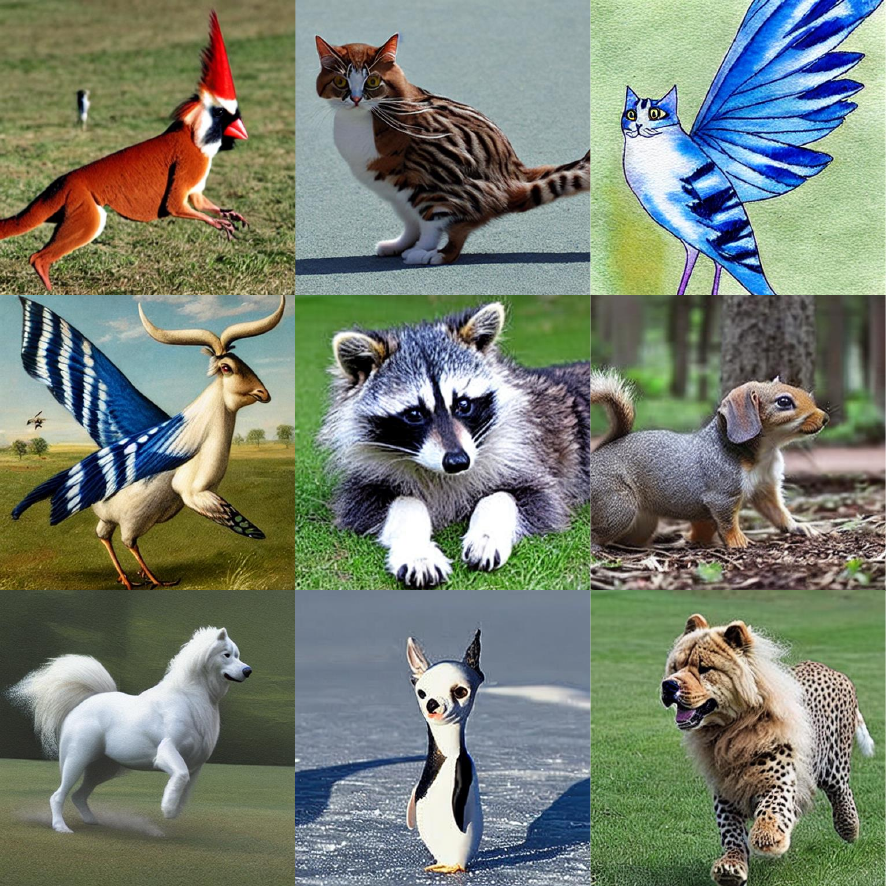}
        \caption{}
    \end{subfigure}
    \begin{subfigure}[b]{\linewidth}
        \includegraphics[width=\linewidth]{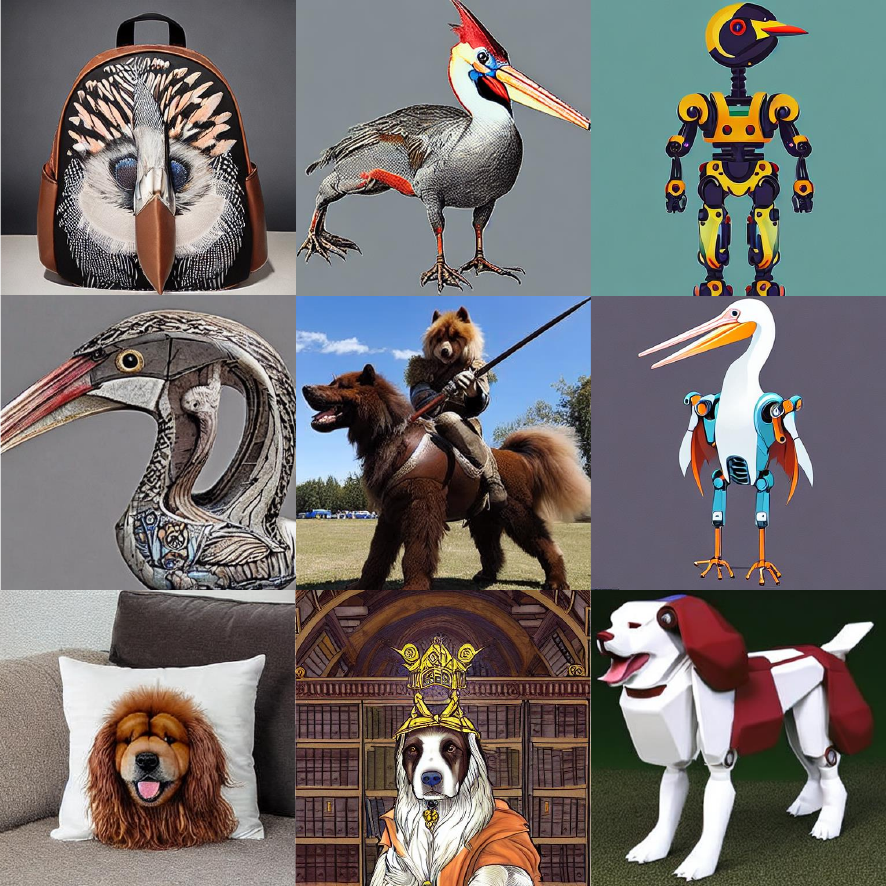}
        \caption{}
    \end{subfigure}

    \caption{We present additional examples of creative generation. \textbf{(a)} displays the effects of transferring learned sub-concepts, \eg, replacing a leopard head with a \texttt{chow}'s head. \textbf{(b)} displays using the learned sub-concepts to inspire some character/product designs.}
    \label{fig:design}
\end{figure}

\clearpage
\section{Further Analysis}

\subsection{Visualization of word embeddings}

\begin{figure}[h]
    \centering
    \begin{subfigure}[b]{0.7\linewidth}
        \includegraphics[width=\linewidth]{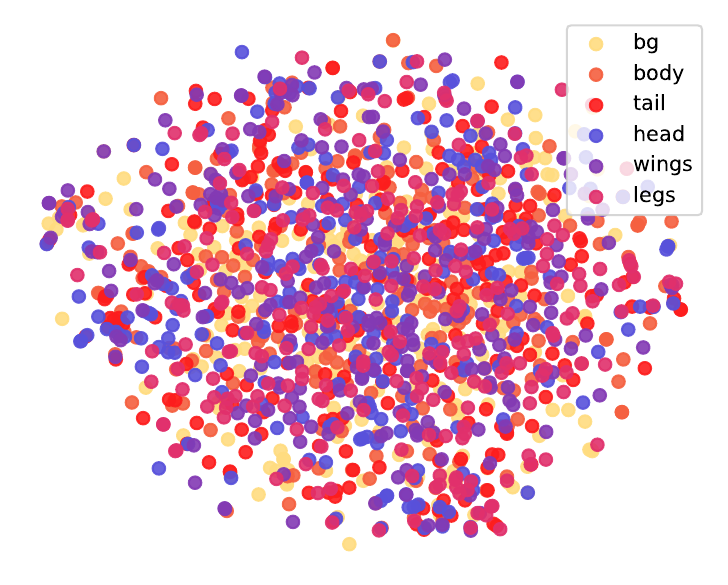}
        \caption{Textual Inversion}
    \end{subfigure}
    \hfill
    \begin{subfigure}[b]{0.7\linewidth}
        \includegraphics[width=\linewidth]{sec/figs/tsne_ti.pdf}
        \caption{Break-a-scene}
    \end{subfigure}
    \hfill
    \begin{subfigure}[b]{0.7\linewidth}
        \includegraphics[width=\linewidth]{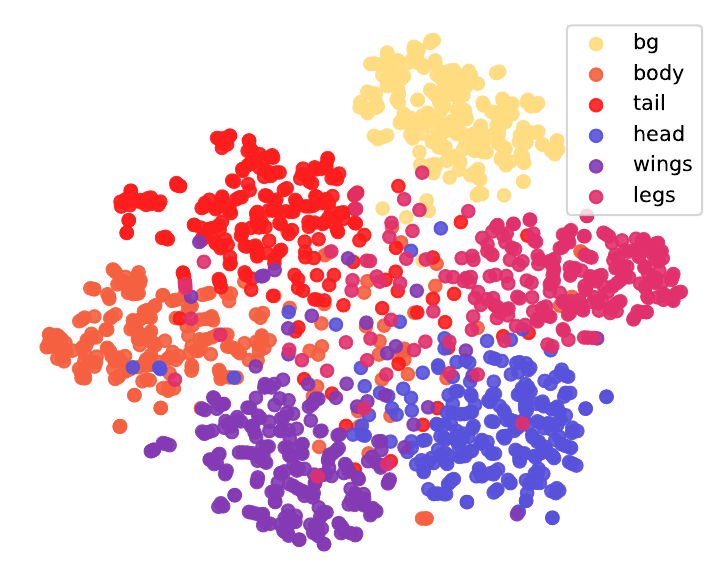}
        \caption{DreamCreature}
    \end{subfigure}
    \caption{2D tSNE \cite{van2008tsne} projection of word embeddings. Different colors represent different body parts (correspond to the segmentation mask in Fig.~\ref{fig:example_seg}).}
    \label{fig:tsne}
\end{figure}

We visualize the word embeddings of learned tokens of Textual Inversion \cite{gal2022textualinversion}, Break-a-scene \cite{avrahami2023breakascene} and our DreamCreature for birds generation (CUB-200-2011 \cite{wah2011cub200}) through tSNE \cite{van2008tsne} in Fig.~\ref{fig:tsne}. In our DreamCreature, the word embeddings are the projected embeddings through Eq. (\textcolor{red}{4})
We can see that our projected version has a better semantic meaning such that the sub-concept embeddings are clustered together by their semantic meaning (\eg, head). We believe this is one of the reasons that our DreamCreature outperforms previous methods in which we can compose all sub-concepts seamlessly yet with higher quality.

\subsection{Attention loss weight}

\begin{figure}[h]
    \centering
    \includegraphics[scale=0.6]{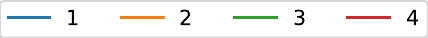} \\ 
    \includegraphics[scale=0.5]{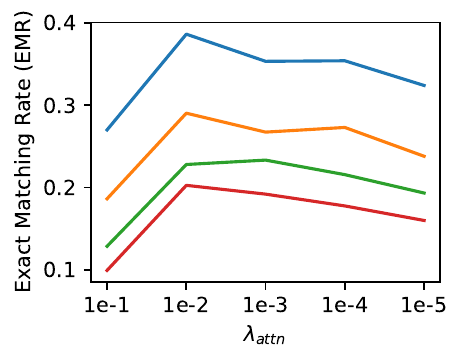}
    \includegraphics[scale=0.5]{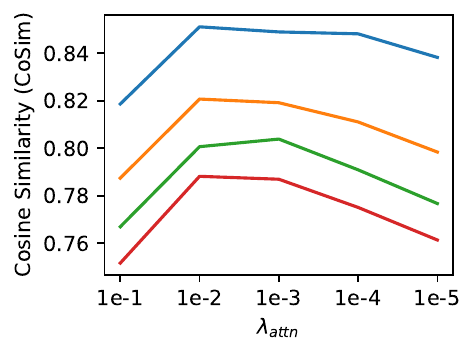}
    \caption{Ablation on the effect of $\lambda_{attn}$ for virtual creature generation on CUB-2011 birds. Different colors represent different numbers of composited sub-concepts.}
    \label{fig:ablattn_vcc}
\end{figure}

In Fig.~\ref{fig:ablattn_vcc} and Tab.~\ref{tab:ablattn}, we summarize the results of our ablation study on the impact of $\lambda_{attn}$. We observed that $\lambda_{attn}=0.01$ frequently yields the best EMR and CoSim scores, while also delivering comparable FID scores. Consequently, we have adopted this value as the default in our experiments.

\begin{table}[h]
    \centering
    \begin{tabular}{c|ccccc}
        $\lambda_{attn}$ & 0.1 & 0.01 & 0.001 & 0.0001 & 0.00001 \\
        \midrule
        FID ($\downarrow$) & 19.08 & 12.86 & 12.44 & 11.78 & \bf 11.64 \\
        EMR ($\uparrow$) & 0.339 & \bf 0.460 & 0.445 & 0.425 & 0.397 \\
        CoSim ($\uparrow$) & 0.851 & \bf 0.882 & 0.880 & 0.878 & 0.872 \\
    \end{tabular}
    \caption{Ablation on the effect of $\lambda_{attn}$ for conventional generation on CUB-200-2011 birds.}
    \label{tab:ablattn}
\end{table}

\subsection{Cross-attention visualization} \label{sec:crossattnvis}

\begin{figure}[h]
    \centering
    \includegraphics[width=\linewidth]{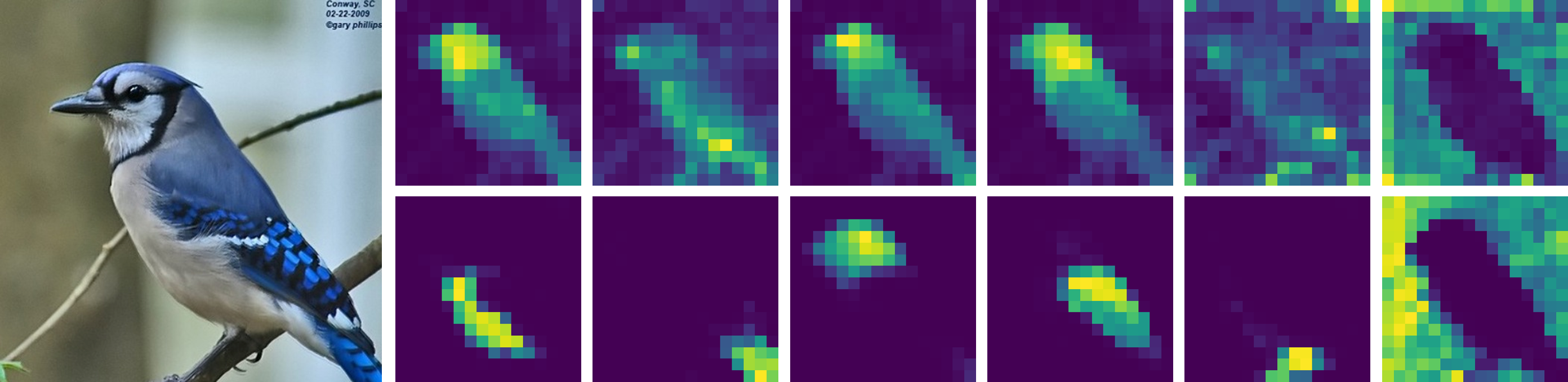}
    \caption{Cross-attention map of each sub-concept (top) without and (bottom) with our attention loss.}
    \label{fig:before_after_attn}
\end{figure}

Our attention loss plays a crucial role in token disentanglement.
We demonstrate the impact of this loss in  Fig.~\ref{fig:before_after_attn},
where we observe significantly enhanced disentanglement after explicitly guiding attention to focus on distinct semantic regions. 

\subsection{Convergence analysis} \label{sec:convergence}

\begin{figure}[h]
    \centering
    \includegraphics[width=\linewidth]{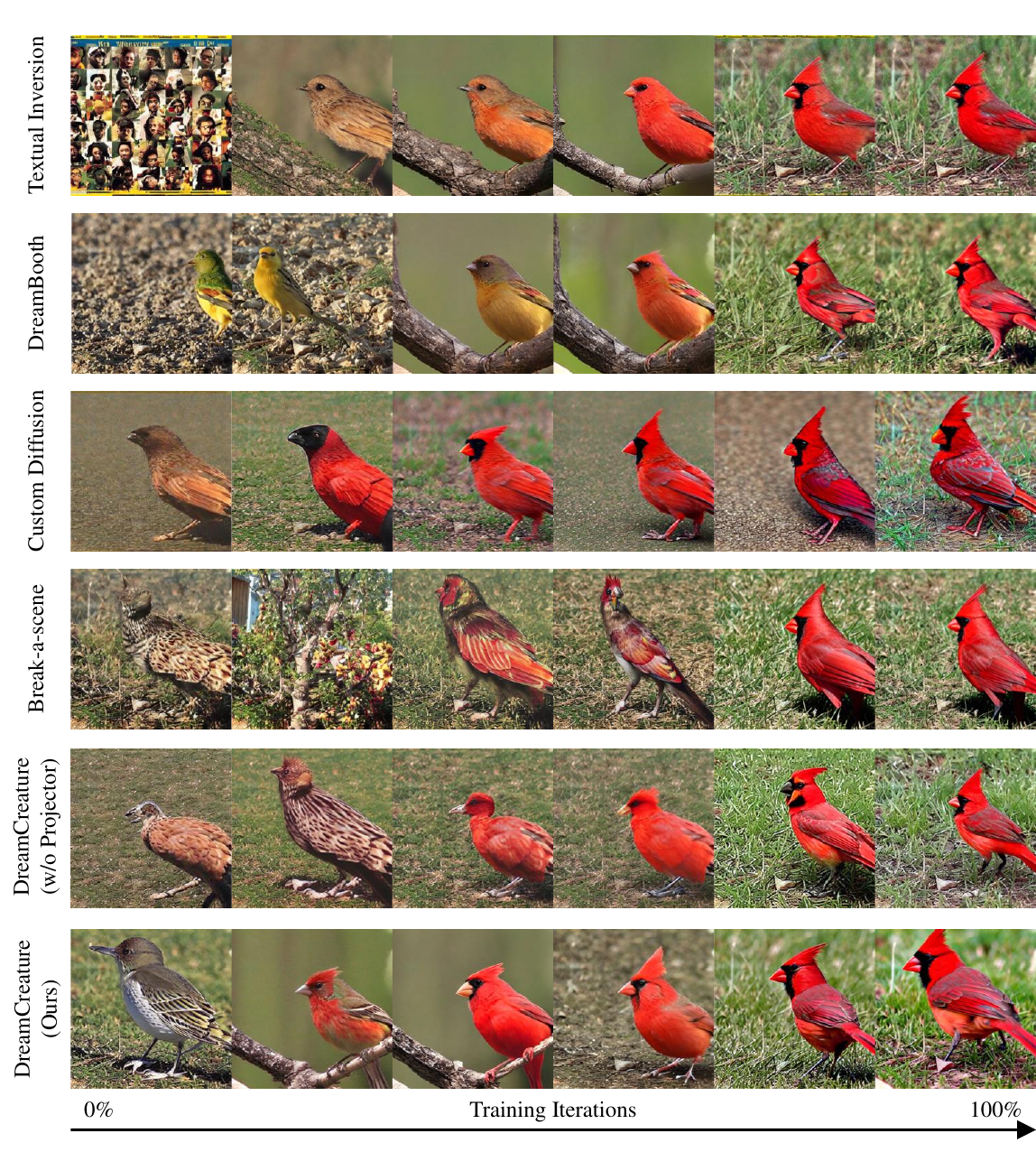} \\  
    \caption{Generated images over different stages of training.}
    \label{fig:training_iters}
\end{figure}

We present a visual comparison of images generated by various methods under the conventional setting in Fig.~\ref{fig:training_iters},
spanning from the initial to the final stages of training. Notably, our DreamCreature demonstrates an ability to learn new concepts at even the early stages of training.

\end{document}